\documentclass[11pt]{article}

\usepackage[preprint]{acl}

\usepackage{times}
\usepackage{latexsym}
\usepackage{booktabs}
\usepackage{multirow}
\usepackage{subcaption}
\usepackage{makecell}
\usepackage{amsmath}
\usepackage[linesnumbered,ruled,vlined]{algorithm2e}

\usepackage[T1]{fontenc}

\usepackage[utf8]{inputenc}

\usepackage{microtype}

\usepackage{inconsolata}

\usepackage{graphicx}

\title{Diversity-Oriented Fine-Tuning for Uncertainty-Based Hallucination Detection}

\author{
  Qiuyuan Li\textsuperscript{1,2}
  \quad
  Hongliang Dai\textsuperscript{1,2}\thanks{Corresponding author.}
  \quad
  Piji Li\textsuperscript{1,2}
  \\[2pt]
  \textsuperscript{1}College of Artificial Intelligence, 
  Nanjing University of Aeronautics and Astronautics, Nanjing, China
  \\
  \textsuperscript{2}The Key Laboratory of Brain-Machine Intelligence Technology, 
  Ministry of Education, Nanjing, China
  \\
  \texttt{\{qiuyli,hongldai,pjli\}@nuaa.edu.cn}
}

\begin{document}
\maketitle
\begin{abstract}


Existing hallucination detection methods are typically conducted at the inference stage, without making any modifications to the model itself. In this paper, we are interested in exploring fine-tuning strategies that enhance the detectability of hallucinations in the resulting model. Focusing on semantic-entropy-based detection, we observe that many erroneous outputs remain undetected because the model produces nearly identical incorrect answers across multiple runs. To address this, we propose diversity-oriented fine-tuning to encourage more varied generations. We introduce two specific strategies: one based on Supervised Fine-Tuning (SFT) and the other on Direct Preference Optimization (DPO). Extensive experiments are conducted to evaluate our approach and analyze the behavior of the models before and after fine-tuning. We find that after adopting our fine-tuning methods, the models become less likely to produce low semantic entropy responses for hallucinated answers, thereby improving the effectiveness of hallucination detection, eventually yielding results better than or comparable with state of the art methods. The code will be publicly released.


\end{abstract}

\begin{figure}[t]
    \centering
    \includegraphics[width=\linewidth]{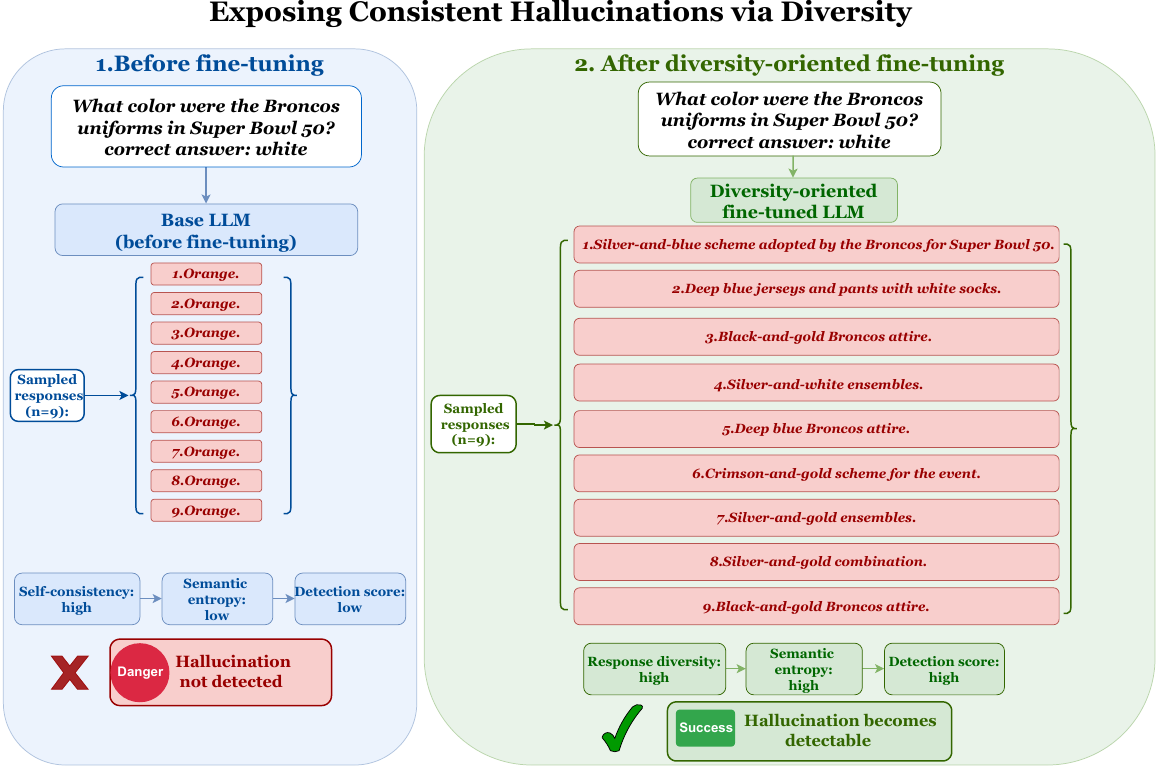}
    \caption{Motivation of diversity-oriented fine-tuning. A base LLM may repeatedly generate the same incorrect answer, yielding low semantic entropy and missed detection. Diversity-oriented fine-tuning induces semantic variation among incorrect samples, increasing entropy and making such hallucinations easier to detect.}
    \label{fig:motivation}
\end{figure}
\textbf{}

\section{Introduction}


Despite significant advancements, Large Language Models (LLMs) remain susceptible to hallucinations: the generation of linguistically plausible yet factually erroneous or unsubstantiated outputs \cite{ji2023survey}. This persistent challenge undermines their reliability and limits their utility in real-world applications \cite{weidinger2021ethical}.
Given the considerable theoretical and practical challenges in fully eliminating hallucinations \cite{xu2024hallucination}, the reliable detection of hallucinatory outputs from LLMs has emerged as a vital research priority \cite{manakul2023selfcheckgpt,kuhn2023semantic,azaria2023internal}.

While many hallucination detection methods have been proposed \cite{ji2023survey,manakul2023selfcheckgpt}, they are typically applied at the inference phase after model training is finalized. Although these post-hoc approaches are easy to deploy, their effectiveness can be limited by the fact that they operate on static models developed without regard for future hallucination detection. Therefore, this work explores fine-tuning strategies that can enhance the detectability of hallucinations in LLMs. 
Specifically, we focus on improving the detection performance of the semantic entropy-based approach \cite{farquhar2024detecting} since it is well recognized. This approach detects hallucinations by identifying semantic inconsistencies across multiple sampled responses.
Our preliminary experiments reveal that when using semantic entropy, a significant amount of hallucinations cannot be detected because the model exhibits high token-level confidence and generates almost identical responses over independent runs.
This observation suggests that diversity-oriented fine-tuning, which encourages the model to generate more varied responses, could potentially trigger semantic inconsistencies for these cases and allow hallucinations to be correctly identified. 

We introduce two diversity-oriented fine-tuning strategies for hallucination detection. In the first approach, Supervised Fine-Tuning (SFT) \cite{ouyang2022training} is applied using a one-to-many mapping, where each input prompt is paired with a set of semantically equivalent target outputs. The second approach leverages Direct Preference Optimization (DPO) \cite{rafailov2023direct}. For a given input, it incentivizes the model to simultaneously produce multiple responses that exhibit high lexical diversity but share the same semantic meaning. Afterwards, we adopt the framework proposed by \citet{farquhar2024detecting} to identify hallucinations at the inference stage by measuring the semantic entropy across the model's multiple responses.

Extensive experiments are conducted to analyze our approach and evaluate its effectiveness. The results demonstrate that both of our fine-tuning strategies enable the model to generate more diverse outputs for a given input and enhance hallucination detection. Our analysis shows that while diversity increases across all instances, the change in semantic consistency \cite{kuhn2023semantic} among the outputs deviates significantly between hallucinating and non-hallucinating cases. Specifically, the fine-tuning amplifies semantic inconsistency in outputs primarily when the model is hallucinating, which effectively  enhances the detection of previously false negative cases through semantic entropy. Consequently, our approach achieves performance that is superior or comparable to current state-of-the-art methods.

Our main contributions as summarized follows:
\begin{itemize}
    \item To the best of our knowledge, we are the first to improve hallucination detection by operating at the fine-tuning stage.
    \item We propose two diversity oriented fine-tuning strategies to improve semantic-entropy-based hallucination detection by addressing cases where hallucinations go undetected because the model generates identical responses across runs. 
    \item We conduct extensive experiments to analyze the behavior of our approach and demonstrate that it consistently boosts hallucination detection performance across diverse datasets.
\end{itemize}

\section{Related Work}

\subsection{Hallucination Detection in Large Language Models}

Hallucination remains a persistent fundamental challenge for large language models, especially in open-domain and knowledge-intensive generation. Existing detection methods can be broadly categorized into three groups. External-verification approaches assess factuality by checking model outputs against retrieved evidence or external knowledge sources \cite{lewis2020retrieval,min2023factscore}. While effective in constrained settings, their performance depends heavily on retrieval quality and external supervision.

Uncertainty-based and consistency-based methods estimate response reliability from token-level likelihoods, log-probabilities, confidence estimates, or agreement across multiple sampled generations \cite{kadavath2022language,lin2022teaching,wang2022self,manakul2023selfcheckgpt}. These methods are attractive because they do not require explicit external evidence, but they can fail when models consistently repeat the same incorrect answer.

Representation-based methods probe hidden states, attention patterns, or activation statistics to identify hallucination-related signals \cite{azaria2023internal,burns2022discovering,su2024unsupervised}. However, these approaches often require auxiliary classifiers or task-specific supervision, increasing system complexity.

\subsection{Generation Diversity in Large Language Models}

Generative diversity has long been studied in natural language generation, including dialogue, summarization, and open-ended text generation. Early approaches mainly rely on inference-time strategies such as temperature scaling, top-$k$, and nucleus sampling \cite{fan2018hierarchical,holtzman2019curious}, which increase surface variation but may not produce meaningful semantic diversity in instruction-following LLMs.

Recent methods further improve diversity through contrastive decoding, which favors tokens preferred by an expert model while penalizing those favored by an interfering model \cite{li2023contrastive,su2022contrastive}. Training-stage objectives, such as SimCTG \cite{su2022contrastive}, Unlikelihood Training \cite{welleck2019neural}, and MMI-based reweighting \cite{li2016diversity}, also encourage diverse outputs, but often introduce additional training or inference complexity.

Data-centric methods such as Mix-Review \cite{he2021analyzing} mainly improve diversity through data augmentation or regularization. However, these methods are not specifically designed to improve uncertainty-based hallucination detection, which is the focus of our work.

\subsection{Semantic Uncertainty and Entropy-Based Measures}


Semantic uncertainty measures variation over meanings rather than surface forms by clustering generated responses into semantic equivalence classes and computing entropy over the resulting distribution \cite{kuhn2023semantic}. It is robust to paraphrasing and suitable for tasks with multiple valid answers \cite{farquhar2024detecting}. However, its effectiveness depends on sufficiently diverse samples: instruction-tuned LLMs may produce limited semantic variation even when uncertain, yielding low or degenerate entropy \cite{sharma2023towards}. Repeated sampling or aggressive decoding can increase diversity, but introduces additional cost and instability \cite{kossen2024semantic}.

\section{Preliminaries and Pilot Study}
\subsection{Preliminaries}

Following existing studies \cite{tong2025halunet,tong2025semantic,wang2025revisiting,hu2025harp}, we focus on hallucination detection for question answering tasks.


Semantic entropy identifies hallucinations by measuring the semantic consistency across multiple independent model generations. For instance, when asked ``\texttt{What is the capital of Canada?}'', if a model produces semantically divergent outputs, such as 
``\texttt{Toronto}'', ``\texttt{Montreal}'', ``\texttt{It's Vancouver}'', and ``\texttt{The answer is Toronto}'', 
the resulting high semantic entropy often indicates a likely hallucination. Conversely, if the model consistently generates semantically equivalent responses, such as ``\texttt{Ottawa}'', ``\texttt{Ottawa is the capital}'', and ``\texttt{It's Ottawa}'', the answer is more likely to be correct.


Based on this observation, this approach groups semantically equivalent answers together and measures uncertainty over meanings rather than token sequences. Specifically, an incremental clustering approach based on Natural Language Inference (NLI) is adopted. Given a question $q$ and a sampled set of candidate answers $A = \{ a_1, a_2, \ldots, a_N \}$, it iterates through each candidate $a_i$ to assign it to a semantic cluster. For each existing cluster $C_j$ represented by $r_j$, it evaluates semantic equivalence using a bi-directional entailment check via an NLI model. A candidate $a_i$ is integrated into cluster $C_j$ if and only if it mutually entails the representative.
If a candidate fails to satisfy this bi-directional condition with all existing clusters, a new cluster is instantiated to accommodate it. Following prior work on semantic uncertainty and semantic entropy \cite{kuhn2023semantic,farquhar2024detecting}, we use incremental semantic clustering based on bi-directional entailment; the complete procedure is summarized in Algorithm \ref{alg:semantic_clustering} for completeness \cite{farquhar2024detecting}.

Finally, the entropy over the distribution of these semantic clusters is calculated. Let $p(C_j)$ denote the empirical proportion of generated sequences assigned to cluster $C_j$. The semantic entropy is defined as:
\[
\mathrm{SE}(q) = - \sum_{C_j} p(C_j) \log_2 p(C_j).
\]
A low SE score reflects strong consensus among the generated outputs, suggesting that most answers convey the same meaning despite surface-level wording differences. Conversely, a high SE score reflects substantial semantic divergence, where the model vacillates between factually incompatible answers, and therefore serves as an uncertainty signal for hallucination detection.

\begin{figure*}[t]
    \centering
    \includegraphics[width=\linewidth]{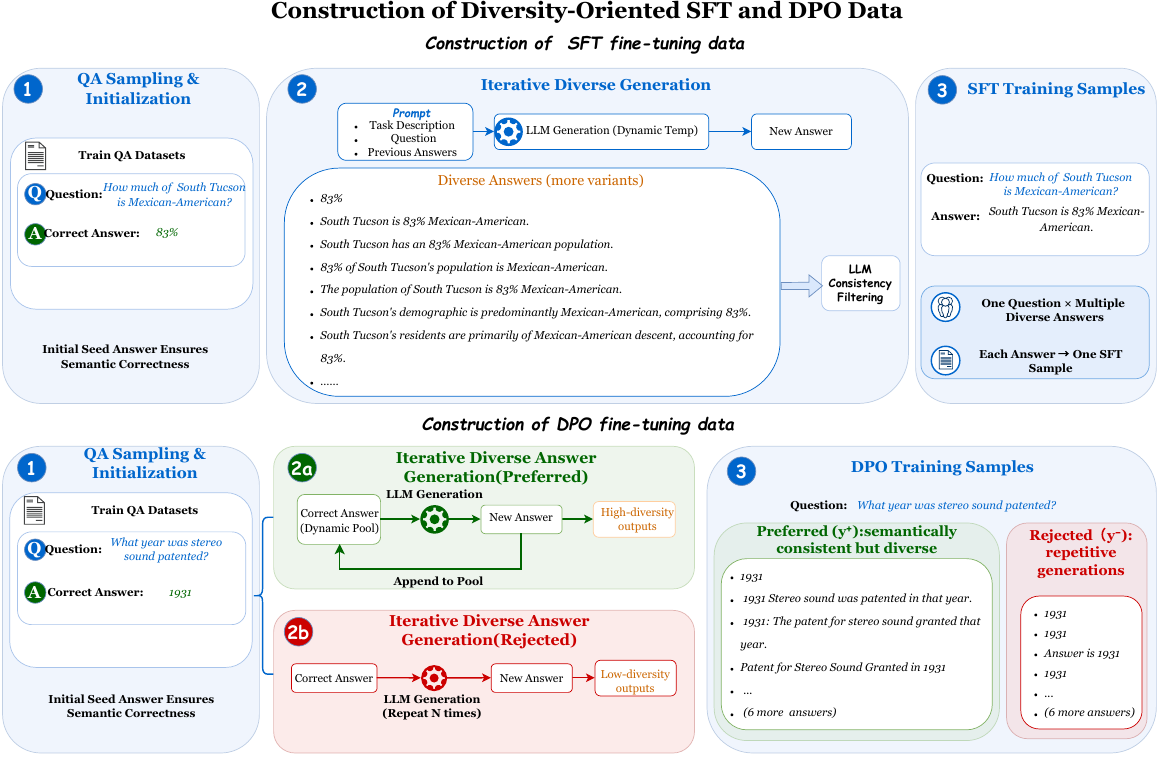}
    \caption{Diversity-oriented fine-tuning data construction. The SFT pipeline generates multiple semantically equivalent answers for each question, creating one training instance per diverse answer. The DPO pipeline constructs preference pairs where high-diversity answer sets are preferred over repetitive low-diversity outputs, encouraging diverse yet semantically faithful generations.}
    \label{fig:data_construction}
\end{figure*}

\subsection{Pilot Study}

By analyzing hallucination detection via semantic entropy, we find that many hallucinations go undetected (false negatives) because the model produces nearly identical responses across multiple independent runs. Specifically, while using LLaMA3-8B-Instruct on the SQuAD dataset to generate 10 independent responses, the model generates identical outputs throughout these iterations for approximately 12\% of hallucinations. This phenomenon is exemplified in Table \ref{tab:case_incorrect} of Appendix \ref{case_study}. Figure~\ref{fig:motivation} illustrates this failure mode. When the base LLM repeatedly generates the same incorrect answer across multiple sampling runs, the sampled responses exhibit high self-consistency and low semantic entropy. As a result, the hallucinated answer is assigned a low uncertainty score and may be missed by consistency-based hallucination detection. This observation motivates diversity-oriented fine-tuning: by encouraging the model to produce more varied responses when it is uncertain, hallucinated cases are more likely to expose semantic inconsistency across samples, leading to higher semantic entropy and improved detectability.

\section{Methodology}


This section introduces our two diversity-oriented fine-tuning strategies: an SFT-based approach which increases the variance of outputs across independent sampling runs; a DPO-based strategy which optimizes the model to generate an output that consists of a diverse set of responses within a single inference step. These two strategies can be integrated into a unified pipeline by performing SFT-based tuning first, followed by DPO.


\subsection{Diversity Fine-tuning with SFT}
\label{sec:sft}



While supervised fine-tuning typically optimizes for labeled input-output pairs, our approach associates each prompt with multiple valid reference responses. This explicitly encourages response diversity at the supervision level.
Formally, we construct the SFT dataset as
\[
D_{\mathrm{SFT}} = \{(x_i, \mathcal{Y}_i)\},
\]
where $x_i$ denotes an input prompt, and $\mathcal{Y}_i = \{y_{i,1}, \ldots, y_{i,K}\}$ represents a set of $K$ semantically equivalent but lexically diverse reference answers corresponding to $x_i$. The SFT objective is defined as the average conditional log-likelihood over all reference answers for each input:
\[
\max_\theta \sum_i \frac{1}{|\mathcal{Y}_i|} \sum_{y \in \mathcal{Y}_i} \log \pi_\theta(y \mid x_i),
\]
where $\pi_\theta(y \mid x)$ denotes the conditional generation probability of the model parameterized by $\theta$.


We construct the diverse SFT dataset using a two-step pipeline: iterative generation via in-context learning and a subsequent verification phase, as illustrated in Figure \ref{fig:data_construction}. 
For each input question, we initialize the process with the original answer and progressively generate multiple variants. Each new generation is conditioned on the previously generated answers, encouraging lexical and structural diversity while ensuring semantic equivalence. 
During the procedure, we increase the generation temperature linearly based on the number of variants produced, further promoting novel expressions. 
After generating the candidate responses, we perform automated semantic consistency verification. Using the LLM as a rigorous judge, we ensure that all variants maintain factual alignment with the original answer. This procedure produces a set of responses per question that are semantically consistent while exhibiting substantial linguistic diversity, making it well suited for diversity-oriented SFT.

\subsection{Diversity Fine-tuning with DPO}

We also design a DPO-based approach inspired by \cite{li2025generating}. 
This method can be applied following our SFT-based strategy, creating a comprehensive diversity-oriented fine-tuning pipeline.


DPO optimizes the model directly on preference data.
Given a human preference dataset
\[
D = \{(x_i, y_{i,1}, y_{i,2})\},
\]
where $y_{i,1}$ represents the preferred response and $y_{i,2}$ represents the non-preferred response, the DPO objective is defined as


\[
\begin{aligned}
\max_{\pi_\theta} \sum_i \log \sigma \biggl( \beta \Bigl[ & \log \frac{\pi_\theta(y_{i,1} \mid x_i)}{\pi_{\text{ref}}(y_{i,1} \mid x_i)} \\
& - \log \frac{\pi_\theta(y_{i,2} \mid x_i)}{\pi_{\text{ref}}(y_{i,2} \mid x_i)} \Bigr] \biggr)
\end{aligned}
\]
where $\pi_\theta(y \mid x)$ denotes the optimized policy, $\pi_{\text{ref}}(y \mid x)$ is a fixed reference policy, $\sigma(\cdot)$ denotes the logistic sigmoid function, and $\beta$ is a temperature parameter that controls the strength of the preference margin. Therefore, DPO fine-tuning relies on well-paired datasets of preferred and non-preferred responses to facilitate contrastive learning in LLMs.





To promote diversity, we ensure a clear structural contrast: preferred outputs consist of multiple distinct responses, whereas non-preferred outputs are composed of multiple repetitive ones.Here, each DPO response is a single concatenated sequence representing a set of answers. The preference is defined over answer sets rather than individual answers: the preferred sequence contains semantically equivalent but lexically diverse answers, whereas the rejected sequence contains repetitive answers.
Figure~\ref{fig:data_construction} shows how we construct training instances for DPO. For preferred outputs, we adopt the same iterative generation procedure used in the SFT-based approach (Section \ref{sec:sft}). After obtaining the diverse set of candidate answers for each question, we combine them as the full output.
For non-preferred outputs, we utilize a static reference strategy. For each question, the model is repeatedly prompted with the ground-truth answer as a fixed reference and instructed to generate another response that is highly similar in wording and structure. Without conditioning on previously generated answers, the model is likely to produce highly homogeneous and repetitive responses. These responses are concatenated to serve as an effective non-preferred output.


\begin{table*}[t]
\centering
\footnotesize
\renewcommand{\arraystretch}{1.15}
\setlength{\tabcolsep}{6pt}
\begin{tabular*}{\textwidth}{@{\extracolsep{\fill}} c l c c c}
\toprule
\multirow{2}{*}{\textbf{Models}} 
& \multirow{2}{*}{\textbf{Fine-Tuning Strategies}} 
& \multicolumn{3}{c}{\textbf{Datasets (AUROC)}} \\
\cmidrule(lr){3-5}
& & \textbf{TriviaQA} & \textbf{SQuAD} & \textbf{NQ-Open} \\
\midrule
\multirow{5}{*}{\textbf{Mistral-7B-Instruct}}
& Baseline            & 0.745 & 0.708 & 0.657 \\
& Standard SFT        & 0.748 & 0.749 & 0.660 \\
& Diverse SFT         & 0.770 & 0.710 & \textbf{0.699} \\
& Diverse DPO         & 0.761 & 0.698 & 0.673 \\
& Diverse SFT + DPO   & \textbf{0.772} & \textbf{0.755} & 0.689 \\
\midrule
\multirow{5}{*}{\textbf{LLaMA3-8B-Instruct}} 
& Baseline            & 0.778 & 0.705 & 0.676 \\
& Standard SFT        & 0.775 & 0.720 & 0.745 \\
& Diverse SFT         & 0.807 & 0.749 & 0.711 \\
& Diverse DPO         & \textbf{0.816} & \textbf{0.759} & 0.746 \\
& Diverse SFT + DPO   & 0.810 & 0.749 & \textbf{0.752} \\
\midrule
\multirow{5}{*}{\textbf{Qwen2.5-7B-Instruct}} 
& Baseline            & 0.732 & 0.692 & 0.715 \\
& Standard SFT        & 0.803 & 0.707 & 0.720 \\
& Diverse SFT         & 0.805 & \textbf{0.748} & \textbf{0.740} \\
& Diverse DPO         & 0.807 & 0.715 & 0.738 \\
& Diverse SFT + DPO   & \textbf{0.839} & 0.747 & 0.739 \\
\bottomrule
\end{tabular*}
\caption{AUROC results of semantic-entropy-based hallucination detection across models, fine-tuning strategies, and datasets. ``Baseline'' directly applies the model without any fine-tuning. The best score for each dataset is highlighted in \textbf{bold}.}
\label{tab:auroc_comparison}
\end{table*}

\section{Experiments}

\subsection{Experimental Setup}

\paragraph{Datasets.}


We evaluate our method on three question answering benchmarks: \textbf{TriviaQA} \cite{joshi2017triviaqa}, which tests general knowledge; \textbf{SQuAD} \cite{rajpurkar2016squad}, a Wikipedia-based reading comprehension dataset; and \textbf{NQ-Open} \cite{kwiatkowski2019natural}, an open-domain subset of Natural Questions consisting of real user queries from Google Search. Their validation sets contain 17,900, 10,600, and 3,610 examples, respectively.

\paragraph{Models.}

Our experiments are conducted across three representative open-source LLM models: \textbf{LLaMA-3-8B-Instruct} \cite{meta2024llama}, \textbf{Mistral-7B-Instruct} \cite{Jiang2023Mistral7} and \textbf{Qwen2.5-7B-instruct} \cite{ahmed2025qwen}. 

\paragraph{Baselines.}

We compare hallucination detection performance with several baselines. Consistency-based methods include \textbf{SelfCheck-BERTScore}, \textbf{SelfCheck-Ngram}, and \textbf{SelfCheck-Prompt} \cite{manakul2023selfcheckgpt}, which estimate hallucination by measuring agreement among multiple sampled responses. Likelihood-based and confidence-based methods include \textbf{Perplexity} and \textbf{P(True)} \cite{kadavath2022language}. Representation-based methods include \textbf{Embedding Regression} \cite{zhang2025icr}, \textbf{ICR Probe} \cite{azaria2023internal}, \textbf{EigenScore} \cite{chen2024inside}, and \textbf{Energy Score}. Similarity-based and entropy-based methods include \textbf{Lexical Similarity}, \textbf{Kernel Language Entropy} \cite{kuhn2023semantic}, and \textbf{Discrete Semantic Entropy} \cite{farquhar2024detecting}. Details of these baselines are provided in Appendix \ref{app:baseline_details}.

\begin{table*}[t]
\centering
\footnotesize
\setlength{\tabcolsep}{6pt}
\renewcommand{\arraystretch}{1.15}

\begin{tabular*}{\textwidth}{@{\extracolsep{\fill}}
>{\centering\arraybackslash}m{2.4cm}
l
c c c 
}
\toprule
\textbf{Models} & \textbf{Methods} 
& \textbf{TriviaQA} & \textbf{SQuAD} & \textbf{NQ-Open}  \\
\midrule

\multirow{9}{*}{\textbf{Mistral-7B}}
& SelfCheck-BERTScore        & 0.675 & 0.633 & 0.625  \\
& SelfCheck-Prompt           & 0.685 & 0.548 & 0.588  \\
& Embedding Regression       & 0.681 & 0.608 & 0.598  \\
& EigenScore                 & 0.758 & 0.712 & 0.664  \\
& Energy Score               & 0.679 & 0.667 & 0.603  \\
& Lexical Similarity         & 0.650 & 0.640 & 0.670  \\
& P(True)                    & 0.724 & 0.619 & 0.674  \\
\cmidrule(lr){2-5}
& \textbf{Ours: SFT}         & 0.770 & 0.710 & \textbf{0.699}  \\
& \textbf{Ours: DPO}         & 0.761 & 0.698 & 0.673  \\
& \textbf{Ours: SFT+DPO}     & \textbf{0.772} & \textbf{0.755} & 0.689  \\
\midrule

\multirow{8}{*}{\textbf{LLaMA3-8B}}
& Perplexity                 & 0.622 & 0.659 & 0.709  \\
& SelfCheck-Ngram            & 0.536 & 0.606 & 0.650  \\
& SelfCheck-Prompt           & 0.758 & 0.630 & 0.690  \\
& EigenScore                 & 0.641 & 0.604 & 0.643  \\
& Energy Score               & 0.691  & 0.588 & 0.645   \\
& Lexical Similarity         & 0.670  & 0.596 & 0.652  \\
& Embedding Regression       & 0.710 & 0.590 & 0.640  \\
& Kernel Language Entropy    & 0.760 & 0.670 & 0.710  \\
\cmidrule(lr){2-5}
& \textbf{Ours: SFT}         & 0.807 & 0.749 & 0.711  \\
& \textbf{Ours: DPO}         & \textbf{0.816} & \textbf{0.759} & 0.746  \\
& \textbf{Ours: SFT+DPO}     & 0.810 & 0.749 & \textbf{0.752}  \\
\midrule

\multirow{6}{*}{\textbf{Qwen2.5-7B}}
& SelfCheck-Prompt           & 0.707 & 0.711 & 0.725  \\
& EigenScore                 & 0.640 & 0.630 & 0.669   \\
& Energy Score               & 0.725  & 0.591  & 0.572   \\
& Lexical Similarity         & 0.652 & 0.532 & 0.686  \\
& ICR Probe                  & 0.758 & 0.745 & \textbf{0.742}     \\
& Discrete Semantic Entropy  & 0.766 & 0.646 & 0.629 \\
\cmidrule(lr){2-5}
& \textbf{Ours: SFT}         & 0.805 & \textbf{0.748} & 0.740  \\
& \textbf{Ours: DPO}         & 0.807 & 0.715 & 0.738  \\
& \textbf{Ours: SFT+DPO}     & \textbf{0.839} & 0.747 & 0.739  \\
\bottomrule
\end{tabular*}

\caption{AUROC comparison grouped by different models and datasets. Within each block, we compare existing hallucination detection methods with our diversity-oriented variants. Section 6.1 offers an analysis of the NQ-Open dataset, examining why our methods yield suboptimal results.}
\label{tab:comparison}
\end{table*}

\paragraph{Evaluation Metrics.}
We evaluate hallucination detection using AUROC. Each instance is treated as positive if the low-temperature answer is incorrect, and the detection score is the semantic entropy computed from multiple sampled responses. The ROC curve is constructed by sweeping the decision threshold over the entropy scores used to classify an answer as hallucinated.

\subsection{Implementation Details}

\paragraph{Answer Generation Setup.}

For each question, we employ a two-stage generation strategy. First, a single response is generated at a low temperature of 0.05 to serve as the correctness reference.
Subsequently, we generate $N$ diverse responses at a high temperature of 0.8. To do this efficiently, the decoder is configured to output $9$ independent sequences in a single forward pass, avoiding repeated calls to the model. 

\paragraph{Fine-tuning Setup.}

We employ a parameter-efficient LoRA strategy for fine-tuning. For SFT fine-tuning, the model is trained for 4 epochs on a maximum of 10000 samples, with a batch size of 2, gradient accumulation steps of 4, and a maximum sequence length of 2048 tokens. Optimization uses the AdamW optimizer with a learning rate of $1e-4$ and a cosine learning rate scheduler, under fp16 precision. For DPO fine-tuning, we adopt a sigmoid preference loss with $\beta = 0.1$ and optimize using AdamW with a learning rate of $2e-6$, also paired with a cosine scheduler. Training runs for 1 epoch with a per-device batch size of 1 and gradient accumulation steps of 4. All DPO experiments are conducted in fp16 precision with a maximum sequence length of 1024 tokens. The prompts used for answer generation, consistency filtering, and data construction are provided in Appendix \ref{app:prompt_details}.

\paragraph{Label Generation Strategy.}

We generate correctness labels with a two-stage verification pipeline. 
We first apply exact matching between the normalized low-temperature answer and the gold answer. If exact matching fails, we use DeBERTa-large-MNLI to determine whether the gold answer entails the low-temperature answer under the question context. The response is labeled as correct if either criterion is satisfied, and incorrect otherwise.

\section{Results and Analysis}
\subsection{Main Results}

Table \ref{tab:auroc_comparison} presents a evaluation of our proposed fine-tuning strategies for hallucination detection. We can see that both of our approaches, Diverse SFT and Diverse DPO, as well as their combined variant (Diverse SFT + DPO), almost always outperforms the un-fine-tuned models (Baseline) and the models that conducts standard SFT. This demonstrates the effectiveness of our methods. Furthermore, we observe that combining Diverse SFT and DPO generally yields intermediate performance levels that fall between those of the two individual methods.

\begin{figure*}[t]
    \centering
    \begin{subfigure}[t]{0.48\textwidth}
        \centering
        \includegraphics[width=\linewidth]{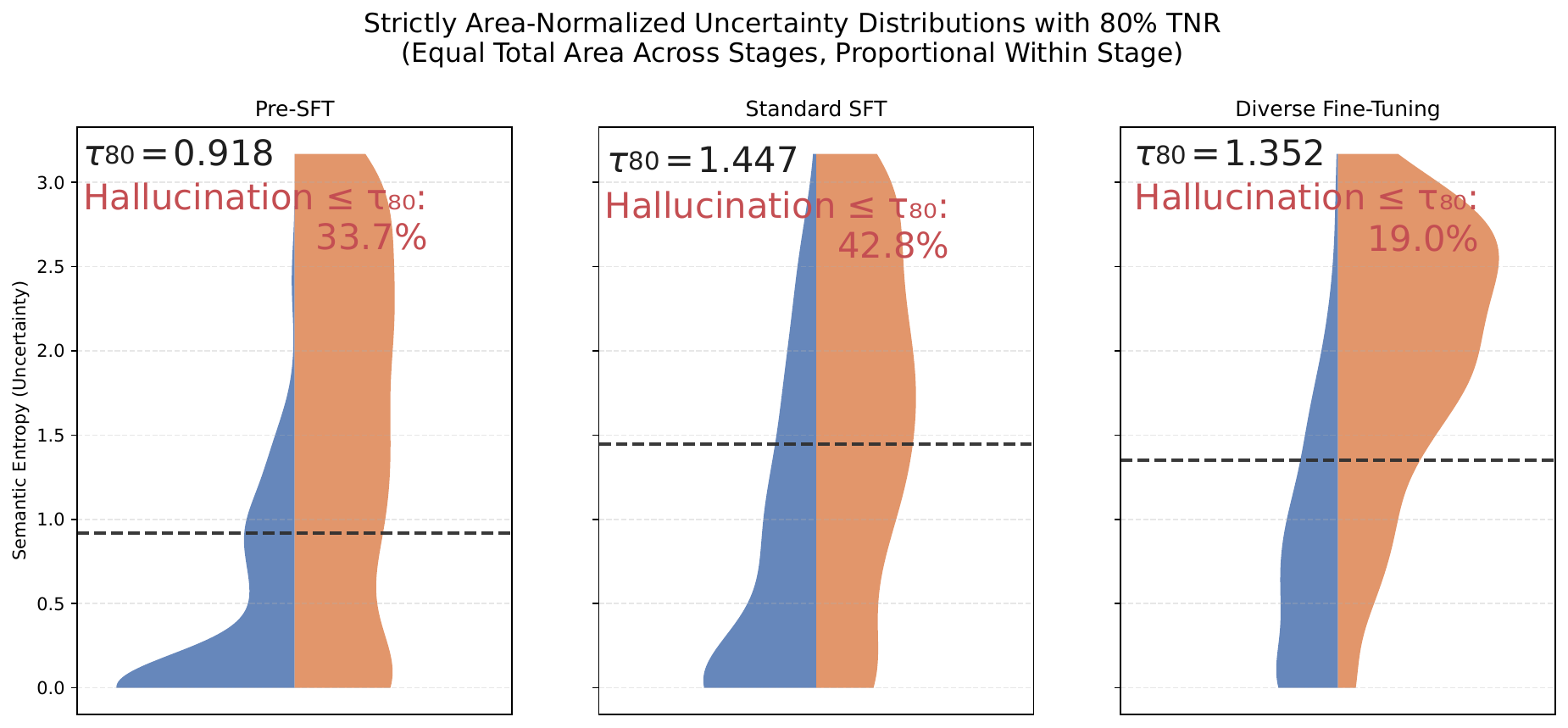}
        \caption{NQ-Open}
        \label{fig:dist_nqopen}
    \end{subfigure}
    \hfill
    \begin{subfigure}[t]{0.48\textwidth}
        \centering
        \includegraphics[width=\linewidth]{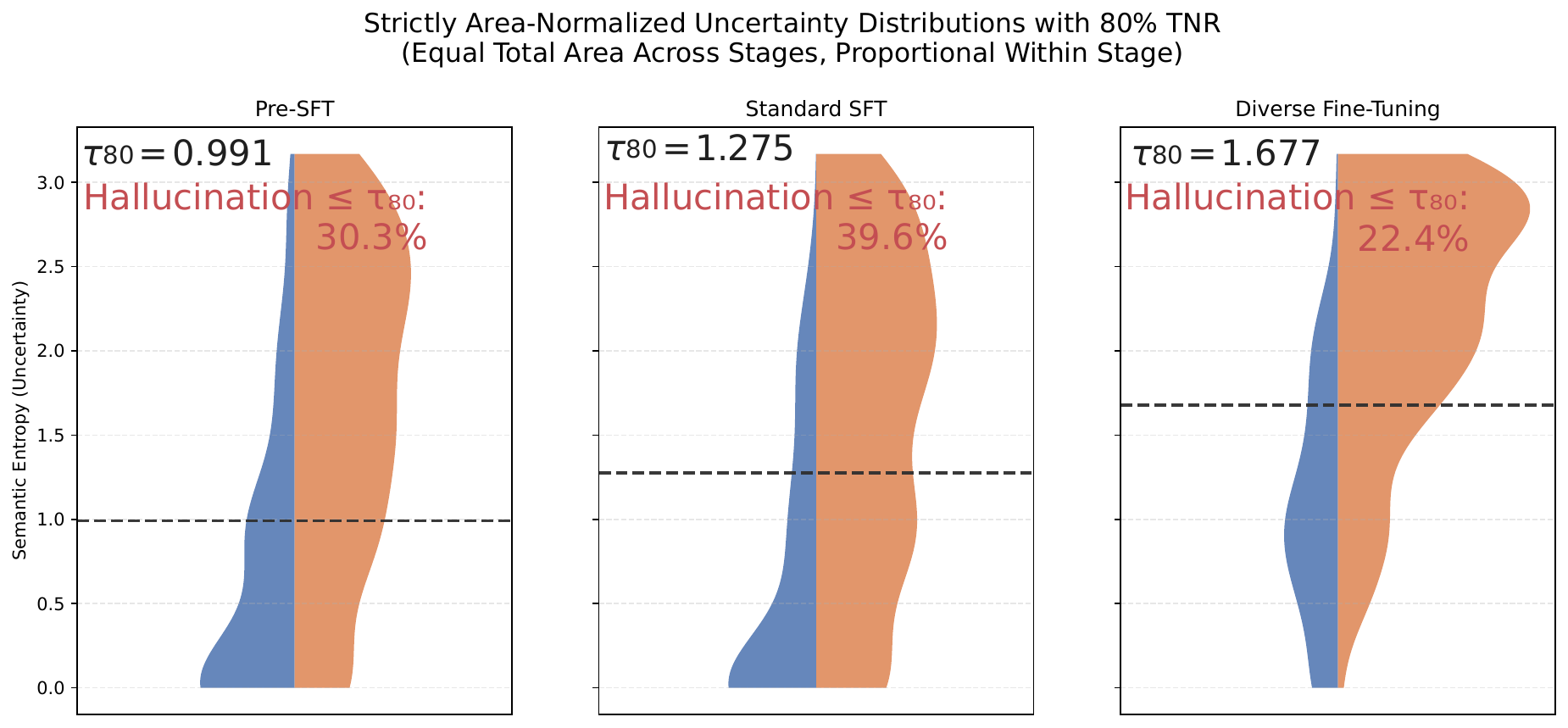}
        \caption{SQuAD}
        \label{fig:dist_squad}
    \end{subfigure}

    \caption{Semantic entropy distributions across fine-tuning stages on NQ-Open and SQuAD. Compared with pre-SFT and standard SFT, diversity-oriented fine-tuning reduces false-negative hallucinations below the 80\% TNR threshold, making hallucinated responses easier to detect.}
    \label{fig:distribution_comparison_all}
\end{figure*}

\begin{table}[t]
\centering
\footnotesize
\setlength{\tabcolsep}{4pt} 
\renewcommand{\arraystretch}{0.95}
\begin{tabular}{l l c c c}
\toprule
\textbf{Models} & \textbf{Methods} & \textbf{TriviaQA} & \textbf{SQuAD} & \textbf{NQ-Open}  \\
\midrule
\multirow{5}{*}{Mistral}
& Baseline  & 0.663 & 0.273 & 0.328  \\
& Std.SFT   & 0.620 & 0.248 & 0.330  \\
& SFT       & 0.628 & 0.300 & 0.338  \\
& SFT+DPO   & 0.615 & 0.298 & 0.340  \\
& DPO       & 0.690 & 0.275 & 0.325  \\
\midrule
\multirow{5}{*}{LLaMA}
& Baseline  & 0.693 & 0.270 & 0.310  \\
& Std.SFT   & 0.555 & 0.215 & 0.268  \\
& SFT       & 0.608 & 0.260 & 0.318  \\
& SFT+DPO   & 0.610 & 0.268 & 0.325  \\
& DPO       & 0.695 & 0.275 & 0.325  \\
\midrule
\multirow{5}{*}{Qwen}
& Baseline  & 0.588 & 0.268 & 0.248 \\
& Std.SFT   & 0.480 & 0.238 & 0.240    \\
& SFT       & 0.518 & 0.263 & 0.238  \\
& SFT+DPO   & 0.528 & 0.238 & 0.245  \\
& DPO       & 0.588 & 0.278 & 0.243  \\
\bottomrule
\end{tabular}
\caption{Answer accuracy of different fine-tuning strategies across models and datasets.}
\label{tab:accuracy_comparison}
\end{table}

Table \ref{tab:comparison} compares our approach with existing hallucination detection methods. Our diversity-enhanced model outperforms prior methods on most datasets. 
However, semantic-entropy-based detection remains less effective for Mistral-7B and LLaMA3-8B on NQ-Open. Since NQ-Open is an open-domain benchmark, a question may have multiple semantically valid answers. In this setting, semantic entropy can split correct but diverse answers into fragmented clusters, producing artificially high uncertainty and degrading detection performance. Representative examples are provided in Appendix~\ref{case_study on Open-Domain QA}.

\subsection{Comparison with Inference-Time Diversity Baselines}
We also compare inference-time diversity baselines, including higher-temperature sampling, nucleus sampling with top-$p=0.95$, and diverse beam search. Since our main baseline already uses high-temperature sampling ($T=0.8$) for semantic entropy computation, these experiments test whether stronger decoding-time diversity can replace fine-tuning. Detailed experimental results are provided in Appendix \ref{app:inference_baselines}, where our fine-tuned models consistently outperform the corresponding inference-time diversity baselines.



\subsection{Analysis of Fine-tuning Effects}

We further analyze how our fine-tuning strategy improves the detectability of hallucinations. Figure \ref{fig:distribution_comparison_all} visualizes the semantic entropy distributions on the NQ-Open and SQuAD datasets before fine-tuning, after standard SFT, and after diverse SFT. We observe that diverse SFT increases semantic entropy overall; however, this increase is more pronounced in hallucinating cases. Specifically, at a threshold yielding an 80\% true negative rate, the proportion of hallucinating cases with semantic entropy below this threshold (False Negatives) drops substantially from 33.7\% to 19.0\% on NQ-Open and from 30.3\% to 22.4\% on SQuAD after diverse SFT. This disproportionate semantic impact on hallucinating versus non-hallucinating cases effectively enhances detectability. The examples in Table \ref{tab:case_correct} and \ref{tab:case_incorrect} of Appendix \ref{case_study} illustrate this phenomenon.


\subsection{Generalizability of Hallucination Detection across Models}

We further examine whether the hallucination signals induced by diversity fine-tuning are model-specific or transferable across architectures. Specifically, we test whether semantic entropy computed from a fine-tuned model can predict hallucinations in responses generated by its un-fine-tuned counterpart. This setting evaluates whether diversity-oriented fine-tuning learns a general uncertainty signal rather than merely overfitting to the fine-tuned model's own generations. If the induced entropy signal remains predictive for base-model outputs, it suggests that the method captures transferable patterns of semantic inconsistency associated with hallucinations. Detailed AUROC results are provided in Appendix \ref{Cross-Model Transferability}.


\subsection{Does Diversity-Oriented Fine-Tuning Increase Hallucinations?}

We evaluate answer accuracy before and after diversity-oriented fine-tuning under the same inference setting. As shown in Table~\ref{tab:accuracy_comparison}, accuracy remains generally stable across datasets, with some model-specific variation. SFT and SFT+DPO reduce accuracy on TriviaQA, while DPO largely preserves or slightly improves it. On SQuAD and NQ-Open, accuracy is comparable to or slightly higher than the baseline, suggesting that diversity-oriented fine-tuning does not substantially harm short-form QA performance. 

\section{Conclusion}

In this work, we propose a diversity-oriented fine-tuning framework that integrates SFT and DPO to make semantic entropy a more reliable signal for hallucination detection. Our method targets cases where a model repeatedly produces the same incorrect answer, causing low uncertainty and false-negative detection. Experiments on three datasets show that our approach achieves performance comparable to or better than state-of-the-art methods. Further analyses indicate that the gains come from making hallucinated cases more distinguishable, rather than merely increasing generation randomness or reducing answer accuracy. We also show that the learned uncertainty signals transfer effectively to base models, demonstrating their practical utility for LLM hallucination detection.

\section*{Limitations}
The proposed diversity-oriented fine-tuning approach improves semantic-entropy-based hallucination detection but has several limitations. First, it relies on high-temperature single-pass decoding to generate multiple distinct responses, which is more efficient than iterative sampling but still incurs additional decoding overhead compared with standard greedy decoding. Second, although our experiments span multiple factual question answering datasets, the evaluation still relies on gold-answer matching and NLI-based semantic equivalence rather than retrieval-based external verification. Extending evaluations to long-form, open-ended, and multi-hop generation tasks with evidence-grounded factuality metrics or retrieval-based verification would provide a more comprehensive assessment of the robustness of our approach.



\bibliography{custom}

@article{lewis2020retrieval,
  title={Retrieval-augmented generation for knowledge-intensive nlp tasks},
  author={Lewis, Patrick and Perez, Ethan and Piktus, Aleksandra and Petroni, Fabio and Karpukhin, Vladimir and Goyal, Naman and K{\"u}ttler, Heinrich and Lewis, Mike and Yih, Wen-tau and Rockt{\"a}schel, Tim and others},
  journal={Advances in neural information processing systems},
  volume={33},
  pages={9459--9474},
  year={2020}
}

@inproceedings{min2023factscore,
  title={Factscore: Fine-grained atomic evaluation of factual precision in long form text generation},
  author={Min, Sewon and Krishna, Kalpesh and Lyu, Xinxi and Lewis, Mike and Yih, Wen-tau and Koh, Pang and Iyyer, Mohit and Zettlemoyer, Luke and Hajishirzi, Hannaneh},
  booktitle={Proceedings of the 2023 Conference on Empirical Methods in Natural Language Processing},
  pages={12076--12100},
  year={2023}
}

@article{kadavath2022language,
  title={Language models (mostly) know what they know},
  author={Kadavath, Saurav and Conerly, Tom and Askell, Amanda and Henighan, Tom and Drain, Dawn and Perez, Ethan and Schiefer, Nicholas and Hatfield-Dodds, Zac and DasSarma, Nova and Tran-Johnson, Eli and others},
  journal={arXiv preprint arXiv:2207.05221},
  year={2022}
}

@article{lin2022teaching,
  title={Teaching models to express their uncertainty in words},
  author={Lin, Stephanie and Hilton, Jacob and Evans, Owain},
  journal={arXiv preprint arXiv:2205.14334},
  year={2022}
}

@article{wang2022self,
  title={Self-consistency improves chain of thought reasoning in language models},
  author={Wang, Xuezhi and Wei, Jason and Schuurmans, Dale and Le, Quoc and Chi, Ed and Narang, Sharan and Chowdhery, Aakanksha and Zhou, Denny},
  journal={arXiv preprint arXiv:2203.11171},
  year={2022}
}

@inproceedings{manakul2023selfcheckgpt,
  title={Selfcheckgpt: Zero-resource black-box hallucination detection for generative large language models},
  author={Manakul, Potsawee and Liusie, Adian and Gales, Mark},
  booktitle={Proceedings of the 2023 conference on empirical methods in natural language processing},
  pages={9004--9017},
  year={2023}
}

@article{azaria2023internal,
  title={The internal state of an LLM knows when it's lying},
  author={Azaria, Amos and Mitchell, Tom},
  journal={arXiv preprint arXiv:2304.13734},
  year={2023}
}

@article{burns2022discovering,
  title={Discovering latent knowledge in language models without supervision},
  author={Burns, Collin and Ye, Haotian and Klein, Dan and Steinhardt, Jacob},
  journal={arXiv preprint arXiv:2212.03827},
  year={2022}
}

@article{su2024unsupervised,
  title={Unsupervised real-time hallucination detection based on the internal states of large language models},
  author={Su, Weihang and Wang, Changyue and Ai, Qingyao and Hu, Yiran and Wu, Zhijing and Zhou, Yujia and Liu, Yiqun},
  journal={arXiv preprint arXiv:2403.06448},
  year={2024}
}

@article{holtzman2019curious,
  title={The curious case of neural text degeneration},
  author={Holtzman, Ari and Buys, Jan and Du, Li and Forbes, Maxwell and Choi, Yejin},
  journal={arXiv preprint arXiv:1904.09751},
  year={2019}
}

@article{fan2018hierarchical,
  title={Hierarchical neural story generation},
  author={Fan, Angela and Lewis, Mike and Dauphin, Yann},
  journal={arXiv preprint arXiv:1805.04833},
  year={2018}
}

@article{ouyang2022training,
  title={Training language models to follow instructions with human feedback},
  author={Ouyang, Long and Wu, Jeffrey and Jiang, Xu and Almeida, Diogo and Wainwright, Carroll and Mishkin, Pamela and Zhang, Chong and Agarwal, Sandhini and Slama, Katarina and Ray, Alex and others},
  journal={Advances in neural information processing systems},
  volume={35},
  pages={27730--27744},
  year={2022}
}

@article{kuhn2023semantic,
  title={Semantic uncertainty: Linguistic invariances for uncertainty estimation in natural language generation},
  author={Kuhn, Lorenz and Gal, Yarin and Farquhar, Sebastian},
  journal={arXiv preprint arXiv:2302.09664},
  year={2023}
}

@article{farquhar2024detecting,
  title={Detecting hallucinations in large language models using semantic entropy},
  author={Farquhar, Sebastian and Kossen, Jannik and Kuhn, Lorenz and Gal, Yarin},
  journal={Nature},
  volume={630},
  number={8017},
  pages={625--630},
  year={2024},
  publisher={Nature Publishing Group UK London}
}

@article{kossen2024semantic,
  title={Semantic entropy probes: Robust and cheap hallucination detection in llms},
  author={Kossen, Jannik and Han, Jiatong and Razzak, Muhammed and Schut, Lisa and Malik, Shreshth and Gal, Yarin},
  journal={arXiv preprint arXiv:2406.15927},
  year={2024}
}

@inproceedings{li2023contrastive,
  title={Contrastive decoding: Open-ended text generation as optimization},
  author={Li, Xiang Lisa and Holtzman, Ari and Fried, Daniel and Liang, Percy and Eisner, Jason and Hashimoto, Tatsunori B and Zettlemoyer, Luke and Lewis, Mike},
  booktitle={Proceedings of the 61st annual meeting of the association for computational linguistics (volume 1: Long papers)},
  pages={12286--12312},
  year={2023}
}

@article{su2022contrastive,
  title={A contrastive framework for neural text generation},
  author={Su, Yixuan and Lan, Tian and Wang, Yan and Yogatama, Dani and Kong, Lingpeng and Collier, Nigel},
  journal={Advances in Neural Information Processing Systems},
  volume={35},
  pages={21548--21561},
  year={2022}
}

@article{welleck2019neural,
  title={Neural text generation with unlikelihood training},
  author={Welleck, Sean and Kulikov, Ilia and Roller, Stephen and Dinan, Emily and Cho, Kyunghyun and Weston, Jason},
  journal={arXiv preprint arXiv:1908.04319},
  year={2019}
}

@inproceedings{li2016diversity,
  title={A diversity-promoting objective function for neural conversation models},
  author={Li, Jiwei and Galley, Michel and Brockett, Chris and Gao, Jianfeng and Dolan, William B},
  booktitle={Proceedings of the 2016 conference of the North American chapter of the association for computational linguistics: human language technologies},
  pages={110--119},
  year={2016}
}

@inproceedings{he2021analyzing,
  title={Analyzing the forgetting problem in pretrain-finetuning of open-domain dialogue response models},
  author={He, Tianxing and Liu, Jun and Cho, Kyunghyun and Ott, Myle and Liu, Bing and Glass, James and Peng, Fuchun},
  booktitle={Proceedings of the 16th Conference of the European Chapter of the Association for Computational Linguistics: Main Volume},
  pages={1121--1133},
  year={2021}
}

@article{sharma2023towards,
  title={Towards understanding sycophancy in language models},
  author={Sharma, Mrinank and Tong, Meg and Korbak, Tomasz and Duvenaud, David and Askell, Amanda and Bowman, Samuel R and Cheng, Newton and Durmus, Esin and Hatfield-Dodds, Zac and Johnston, Scott R and others},
  journal={arXiv preprint arXiv:2310.13548},
  year={2023}
}

@article{wang2025revisiting,
  title={Revisiting Hallucination Detection with Effective Rank-based Uncertainty},
  author={Wang, Rui and Wei, Zeming and Yue, Guanzhang and Sun, Meng},
  journal={arXiv preprint arXiv:2510.08389},
  year={2025}
}

@article{chen2024inside,
  title={INSIDE: LLMs' internal states retain the power of hallucination detection},
  author={Chen, Chao and Liu, Kai and Chen, Ze and Gu, Yi and Wu, Yue and Tao, Mingyuan and Fu, Zhihang and Ye, Jieping},
  journal={arXiv preprint arXiv:2402.03744},
  year={2024}
}

@inproceedings{zhang2025icr,
  title={ICR probe: Tracking hidden state dynamics for reliable hallucination detection in LLMs},
  author={Zhang, Zhenliang and Hu, Xinyu and Zhang, Huixuan and Zhang, Junzhe and Wan, Xiaojun},
  booktitle={Proceedings of the 63rd Annual Meeting of the Association for Computational Linguistics (Volume 1: Long Papers)},
  pages={17986--18002},
  year={2025}
}

@article{meta2024llama,
  title={The llama 3 herd of models},
  author={Meta, AI and Jauhri, Abhinav and Pandey, Abhinav and Kadian, Abhishek and Al-Dahle, Ahmad and Letman, Aiesha and Mathur, Akhil and Schelten, Alan and Yang, Amy and Fan, Angela and others},
  journal={arXiv preprint arXiv:2407.21783},
  volume={2},
  year={2024}
}

@article{ahmed2025qwen,
  title={Qwen 2.5: A comprehensive review of the leading resource-efficient llm with potentioal to surpass all competitors},
  author={Ahmed, Imtiaz and Islam, Sadman and Datta, Partha Protim and Kabir, Imran and Chowdhury, Naseef Ur Rahman and Haque, Ahshanul},
  journal={Authorea Preprints},
  year={2025},
  publisher={Authorea}
}

@article{rajpurkar2016squad,
  title={Squad: 100,000+ questions for machine comprehension of text},
  author={Rajpurkar, Pranav and Zhang, Jian and Lopyrev, Konstantin and Liang, Percy},
  journal={arXiv preprint arXiv:1606.05250},
  year={2016}
}

@article{joshi2017triviaqa,
  title={Triviaqa: A large scale distantly supervised challenge dataset for reading comprehension},
  author={Joshi, Mandar and Choi, Eunsol and Weld, Daniel S and Zettlemoyer, Luke},
  journal={arXiv preprint arXiv:1705.03551},
  year={2017}
}

@article{kwiatkowski2019natural,
  title={Natural questions: a benchmark for question answering research},
  author={Kwiatkowski, Tom and Palomaki, Jennimaria and Redfield, Olivia and Collins, Michael and Parikh, Ankur and Alberti, Chris and Epstein, Danielle and Polosukhin, Illia and Devlin, Jacob and Lee, Kenton and others},
  journal={Transactions of the Association for Computational Linguistics},
  volume={7},
  pages={453--466},
  year={2019},
  publisher={MIT Press One Rogers Street, Cambridge, MA 02142-1209, USA journals-info~…}
}

@article{Jiang2023Mistral7,
  title={Mistral 7B},
  author={Albert Qiaochu Jiang and Alexandre Sablayrolles and Arthur Mensch and Chris Bamford and Devendra Singh Chaplot and Diego de Las Casas and Florian Bressand and Gianna Lengyel and Guillaume Lample and Lucile Saulnier and L{\'e}lio Renard Lavaud and Marie-Anne Lachaux and Pierre Stock and Teven Le Scao and Thibaut Lavril and Thomas Wang and Timoth{\'e}e Lacroix and William El Sayed},
  journal={ArXiv},
  year={2023},
  volume={abs/2310.06825},
  url={https://api.semanticscholar.org/CorpusID:263830494}
}

@article{ji2023survey,
  title={Survey of hallucination in natural language generation},
  author={Ji, Ziwei and Lee, Nayeon and Frieske, Rita and Yu, Tiezheng and Su, Dan and Xu, Yan and Ishii, Etsuko and Bang, Ye Jin and Madotto, Andrea and Fung, Pascale},
  journal={ACM computing surveys},
  volume={55},
  number={12},
  pages={1--38},
  year={2023},
  publisher={ACM New York, NY}
}

@article{rafailov2023direct,
  title={Direct preference optimization: Your language model is secretly a reward model},
  author={Rafailov, Rafael and Sharma, Archit and Mitchell, Eric and Manning, Christopher D and Ermon, Stefano and Finn, Chelsea},
  journal={Advances in neural information processing systems},
  volume={36},
  pages={53728--53741},
  year={2023}
}

@article{li2025generating,
  title={Generating Diverse Training Samples for Relation Extraction with Large Language Models},
  author={Li, Zexuan and Dai, Hongliang and Li, Piji},
  journal={arXiv preprint arXiv:2505.23108},
  year={2025}
}

@article{weidinger2021ethical,
  title={Ethical and social risks of harm from language models},
  author={Weidinger, Laura and Mellor, John and Rauh, Maribeth and Griffin, Conor and Uesato, Jonathan and Huang, Po-Sen and Cheng, Myra and Glaese, Mia and Balle, Borja and Kasirzadeh, Atoosa and others},
  journal={arXiv preprint arXiv:2112.04359},
  year={2021}
}

@article{xu2024hallucination,
  title={Hallucination is inevitable: An innate limitation of large language models},
  author={Xu, Ziwei and Jain, Sanjay and Kankanhalli, Mohan},
  journal={arXiv preprint arXiv:2401.11817},
  year={2024}
}

@article{tong2025halunet,
  title={HaluNet: Multi-Granular Uncertainty Modeling for Efficient Hallucination Detection in LLM Question Answering},
  author={Tong, Chaodong and Zhang, Qi and Gao, Jiayang and Jiang, Lei and Liu, Yanbing and Sun, Nannan},
  journal={arXiv preprint arXiv:2512.24562},
  year={2025}
}

@article{tong2025semantic,
  title={Semantic Reformulation Entropy for Robust Hallucination Detection in QA Tasks},
  author={Tong, Chaodong and Zhang, Qi and Jiang, Lei and Liu, Yanbing and Sun, Nannan and Li, Wei},
  journal={arXiv preprint arXiv:2509.17445},
  year={2025}
}

@article{hu2025harp,
  title={HARP: Hallucination Detection via Reasoning Subspace Projection},
  author={Hu, Junjie and Tu, Gang and Cheng, ShengYu and Li, Jinxin and Wang, Jinting and Chen, Rui and Zhou, Zhilong and Shan, Dongbo},
  journal={arXiv preprint arXiv:2509.11536},
  year={2025}
}

\appendix
\label{sec:appendix}

\section{Experimental Details}

\subsection{Semantic Clustering Algorithm}

We compute semantic entropy by clustering generated responses into semantically equivalent groups using an incremental bi-directional entailment criterion. Each answer is sequentially matched to an existing cluster based on mutual entailment with a cluster representative, otherwise, a new cluster is created. Semantic entropy is then calculated from the resulting cluster size distribution, where higher entropy indicates greater semantic uncertainty. Algorithm \ref{alg:semantic_clustering} details the clustering procedure.

\begin{algorithm}[t]
\caption{Incremental Semantic Clustering via Bi-directional Entailment}
\label{alg:semantic_clustering}
\SetAlgoLined
\textbf{Input:} Question $q$, List of generated answers $\mathcal{A} = \{a_1, \dots, a_N\}$ \\
\textbf{Output:} Semantic Entropy $H$ \\
Initialize clusters $\mathcal{C} \leftarrow \emptyset$ \\
\For{$i \leftarrow 1$ \KwTo $N$}{
    $a_i \leftarrow \text{Preprocess}(a_i)$ \\
    $\textit{matched} \leftarrow \text{False}$ \\
    \For{$C_k \in \mathcal{C}$}{
        $r_k \leftarrow C_k[0]$ \\
        \If{$\text{BiDirectionalEntailment}(q, a_i, r_k)$}{
            $C_k \leftarrow C_k \cup \{a_i\}$ \\
            $\textit{matched} \leftarrow \text{True}$ \\
            \textbf{break}
        }
    }
    \If{\textbf{not} $\textit{matched}$}{
        Create new cluster $C_{\text{new}} \leftarrow \{a_i\}$ \\
        $\mathcal{C} \leftarrow \mathcal{C} \cup \{C_{\text{new}}\}$
    }
}
Compute probability distribution $P(C_k) = |C_k| / N$ \\
Compute Entropy $H = -\sum P(C_k) \log_2 P(C_k)$ \\
\Return $H$
\end{algorithm}

\subsection{Baseline Methods}
\label{app:baseline_details}

This section briefly describes the hallucination detection baselines used in our experiments. These methods cover different uncertainty signals, including generation consistency, likelihood-based confidence, representation geometry, and semantic uncertainty.

\paragraph{SelfCheck Methods.}
SelfCheck-GPT detects hallucinations by measuring consistency across multiple sampled responses. We include SelfCheck variants based on n-gram overlap, BERTScore similarity, and prompt-based consistency checking. These methods assume that factual answers are more consistent across generations, whereas hallucinated answers tend to show lower agreement.

\paragraph{Perplexity.}
Perplexity uses token-level likelihood as a confidence signal. Responses with lower likelihood or higher perplexity are treated as more uncertain and therefore more likely to be hallucinated.

\paragraph{P(True).}
The P(True) estimator asks the model to estimate the probability that its generated answer is correct. A lower estimated probability indicates higher hallucination risk.

\paragraph{Embedding Regression.}
Embedding regression trains a lightweight predictor over hidden representations to estimate answer correctness. It assumes that factual and hallucinated responses are separable in the model's representation space.

\paragraph{ICR Probe.}
ICR Probe detects hallucinations from internal model states. It uses layer-wise hidden-state dynamics to train a probe for distinguishing correct and incorrect generations.

\paragraph{EigenScore.}
EigenScore measures uncertainty through the geometry of hidden representations. It computes dispersion statistics from hidden-state embeddings across multiple generations, where larger representational divergence indicates higher uncertainty.

\paragraph{Energy Score.}
Energy Score measures uncertainty from the token-level predictive distribution. We compute it from the logits along the decoding trajectory, where a larger energy value indicates lower model confidence and higher hallucination risk.

\paragraph{Lexical Similarity.}
Lexical Similarity measures surface-form agreement among sampled responses. We compute the average lexical overlap across generations and use lower similarity as an indicator of higher uncertainty.

\paragraph{Kernel Language Entropy.}
Kernel Language Entropy estimates uncertainty by computing entropy in a kernel-induced space over multiple generations, capturing distributional variation among generated responses.

\paragraph{Discrete Semantic Entropy.}
Discrete Semantic Entropy clusters generated answers into semantic equivalence classes and computes Shannon entropy over the resulting cluster distribution. Unlike token-level entropy, it measures uncertainty over meanings rather than surface forms.

\subsection{Prompt and Data Generation Details}
\label{app:prompt_details}

To improve reproducibility, we describe the prompts and decoding parameters used to construct SFT and DPO data. Our data construction follows an iterative generation and verification pipeline. The generation prompt include the question and previously generated answers, instructing the model to produce one new semantically consistent paraphrase at each iteration. The generated variants are then verified by an LLM-based consistency judge, and inconsistent answer sets are discarded.

\paragraph{Prompt for answer generation.}
For answer generation with the fine-tuned model, we use the following prompt. 

\begin{quote}
\small
\texttt{You are a helpful assistant. Use plain text phrases or sentences to provide answers.}
\end{quote}

\paragraph{Iterative Diverse Answer Generation for SFT.}

For each question, we start from the original reference answer and iteratively generate additional answer variants. At each step, the model is conditioned on the question and the previously generated answers, and the model is instructed to produce one new answer that is semantically consistent but phrased differently.

\begin{quote}
\small
\texttt{
Task: Answer the following question based on previous examples. \\
Question: \{question\} \\
Previous answers: \{existing\_examples\} \\
Instruction: Generate ONE new answer that is semantically correct and consistent with the examples above, but phrased differently. Try to use different sentence structures, synonyms, or formats compared to the previous answers. \\
IMPORTANT: Only output the answer itself.
}
\end{quote}

To encourage lexical diversity, we use a dynamic temperature schedule. 
As more variants are generated, the decoding temperature is linearly increased from 0.7 to 1.1. This encourages later generations to explore more diverse surface forms while remaining conditioned on the previous valid examples.

\paragraph{LLM-based Consistency Filtering.}

After generating the answer variants for a question, we apply an LLM-based consistency check to verify that the generated answers are semantically equivalent and factually consistent. The judge evaluates the entire answer set for one question rather than each answer independently.

\begin{quote}
\small
\texttt{
You are a strict logic consistency judge. Here is a question and a list of generated answers. \\
Question: \{question\} \\
Generated Answers: \{answers\_block\} \\
Task: Determine if ALL the answers above represent the same semantic meaning and are factually consistent with each other. Minor phrasing differences (e.g., `Steve Jobs' vs `Steven Paul Jobs') are acceptable, but factual contradictions or entirely different entities are NOT allowed. Please respond with exactly `YES' if they are consistent, or `NO' if they are not. Do not provide explanations. 
}
\end{quote}

The filtering decision is binary. 
If the judge outputs \texttt{NO}, the entire generated answer set is discarded. 
This filtering step ensures that the SFT data increases answer diversity without introducing factual inconsistency.

\paragraph{DPO Preference Data Construction.}

For DPO training, we construct preference pairs that contrast diverse answer sets with repetitive answer sets. Given a question and its reference answers, we use the first reference answer as the seed answer. The preferred response is constructed by iteratively generating diverse paraphrases conditioned on all previously generated variants. The rejected response is constructed by repeatedly prompting the model with only the seed answer and asking for an almost identical reformulation. Thus, both responses are intended to preserve the same factual content, but the preferred response contains higher lexical and structural diversity, while the rejected response is more repetitive. 

\paragraph{Diverse Prompt for Preferred Responses.}
For constructing preferred responses in DPO training, we use the following prompt to generate semantically consistent but lexically diverse answer variants.

\begin{quote}
\small
\texttt{
Question: \{question\} \\
\\
Previous answers: \{existing\_examples\} \\
\\
Instruction: Generate ONE new answer that conveys the SAME meaning as the answers above, remaining fully consistent with their factual content. Rephrase it naturally using different wording, structure, or tone to ensure diversity in expression, but DO NOT change or add any information. \\
\\
Output rules: \\
- Output ONLY the new answer text (no `Answer:', `Response:', numbering, quotes, or explanations). \\
- The answer must stay semantically identical to the given examples. \\
- Focus on paraphrasing for variety---change phrasing, not meaning. \\
- Keep it concise, fluent, and in the same style as the examples.
}
\end{quote}

\paragraph{Repetitive Prompt for Rejected Responses.}
For constructing rejected responses in DPO training, we use the following prompt to generate repetitive answer variants that preserve the same meaning but differ only minimally in wording.

\begin{quote}
\small
\texttt{
Question: \{question\} \\
\\
Existing answers: \{seed\_answer\} \\
\\
Instruction: Generate ONE answer that is semantically correct and consistent with the examples above, but phrased almost identically, with very minor variations. The answer should be the same in content and meaning, but with small differences in wording or structure. Only output the answer itself.
}
\end{quote}

\paragraph{Hyperparameters.}

As shown in Table~\ref{tab:data_generation_hyperparams}, we use the following parameters during data generation.

\begin{table}[t]
\centering
\footnotesize
\setlength{\tabcolsep}{3pt}
\renewcommand{\arraystretch}{1.08}
\begin{tabular*}{\columnwidth}{@{\extracolsep{\fill}}lc@{}}
\hline
\textbf{Parameter} & \textbf{Value} \\
\hline
Variants per question & 9 \\
Max. DPO samples & 1000 \\
Max. generation tokens & 100 \\
Decoding strategy & nucleus sampling \\
SFT top-$p$ & 0.95 \\
DPO top-$p$ & 0.90 \\
SFT temperature & 0.7--1.1 \\
DPO temperature & 0.7 \\
\hline
\end{tabular*}
\caption{Hyperparameters used for SFT and DPO data generation.}
\label{tab:data_generation_hyperparams}
\end{table}

\section{Additional Results}
\subsection{Cross-Model Transferability Analysis} 
\label{Cross-Model Transferability}
To examine whether hallucination signals induced by diversity fine-tuning are model-specific or transferable, we report cross-model hallucination detection results in Table~\ref{tab:transfer_auc}. The results show that the method exhibits consistent transferability across multiple datasets and base models, indicating that diversity fine-tuning induces more generalized uncertainty signals rather than overfitting to a specific model.

\begin{figure*}[t]
    \centering

    \begin{subfigure}[t]{0.32\textwidth}
        \centering
        \includegraphics[width=\linewidth]{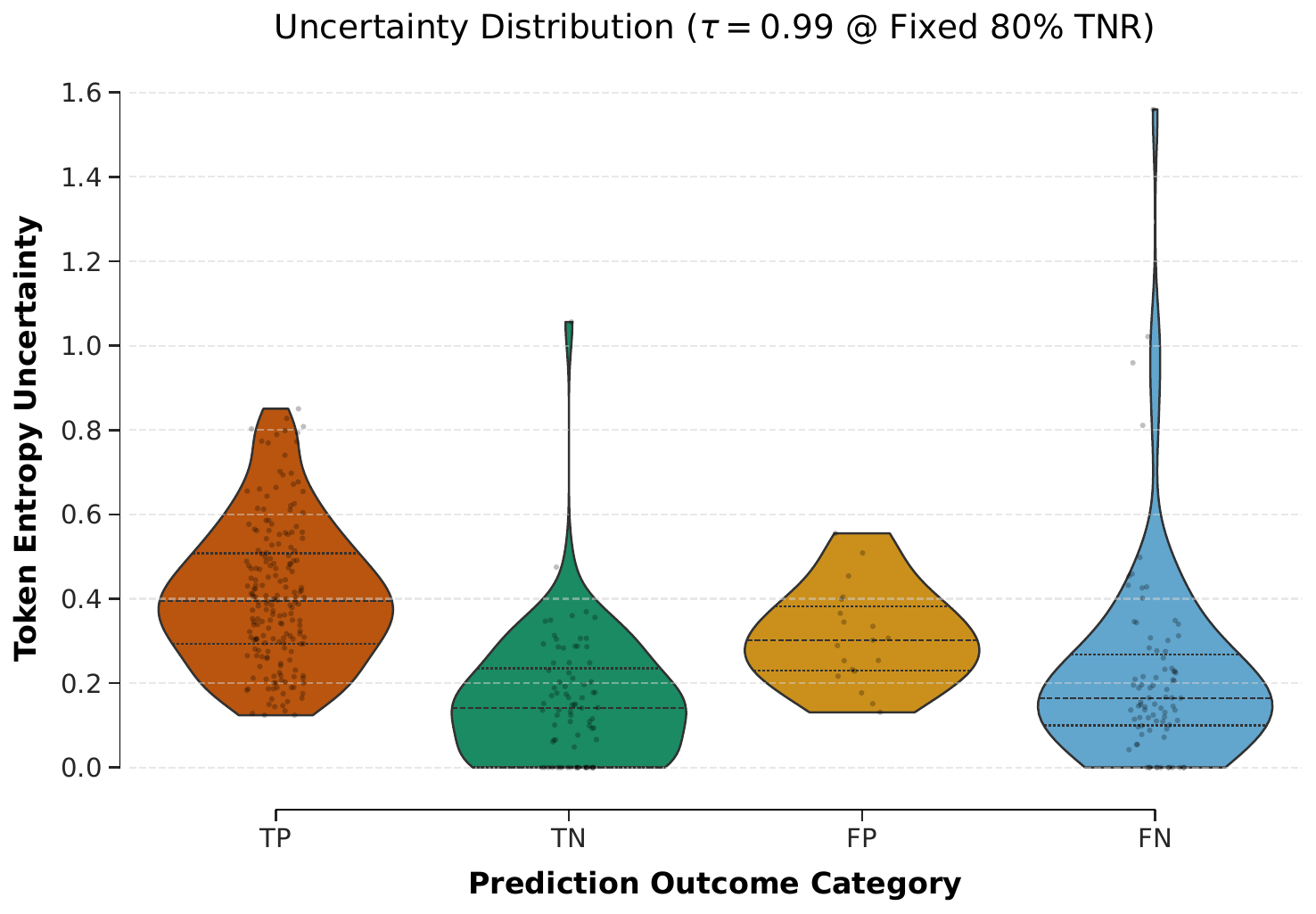}
        \caption{Before Fine-Tuning}
        \label{fig:squad_dist_before}
    \end{subfigure}
    \hfill
    \begin{subfigure}[t]{0.32\textwidth}
        \centering
        \includegraphics[width=\linewidth]{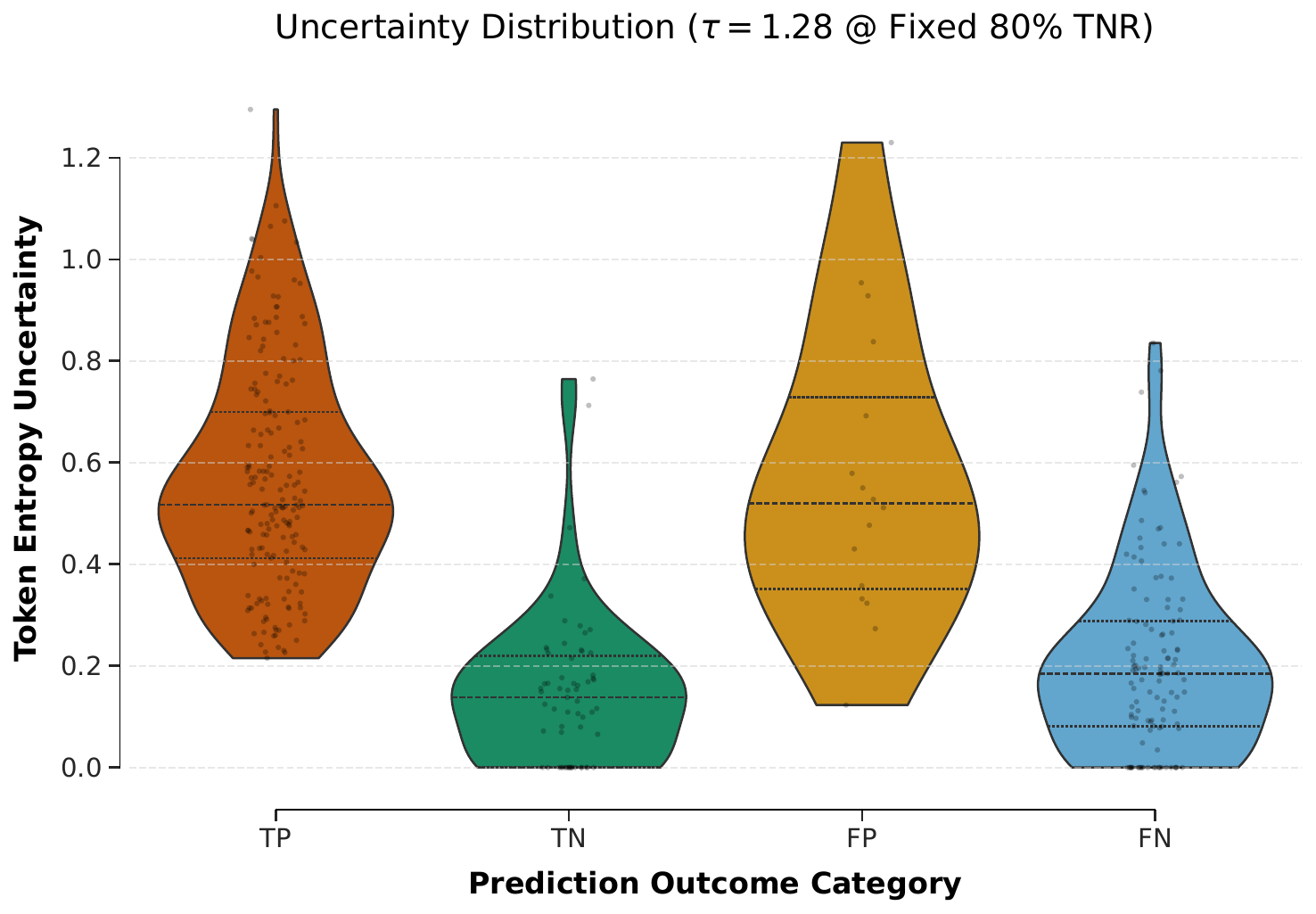}
        \caption{After Standard SFT}
        \label{fig:squad_dist_sft}
    \end{subfigure}
    \hfill
    \begin{subfigure}[t]{0.32\textwidth}
        \centering
        \includegraphics[width=\linewidth]{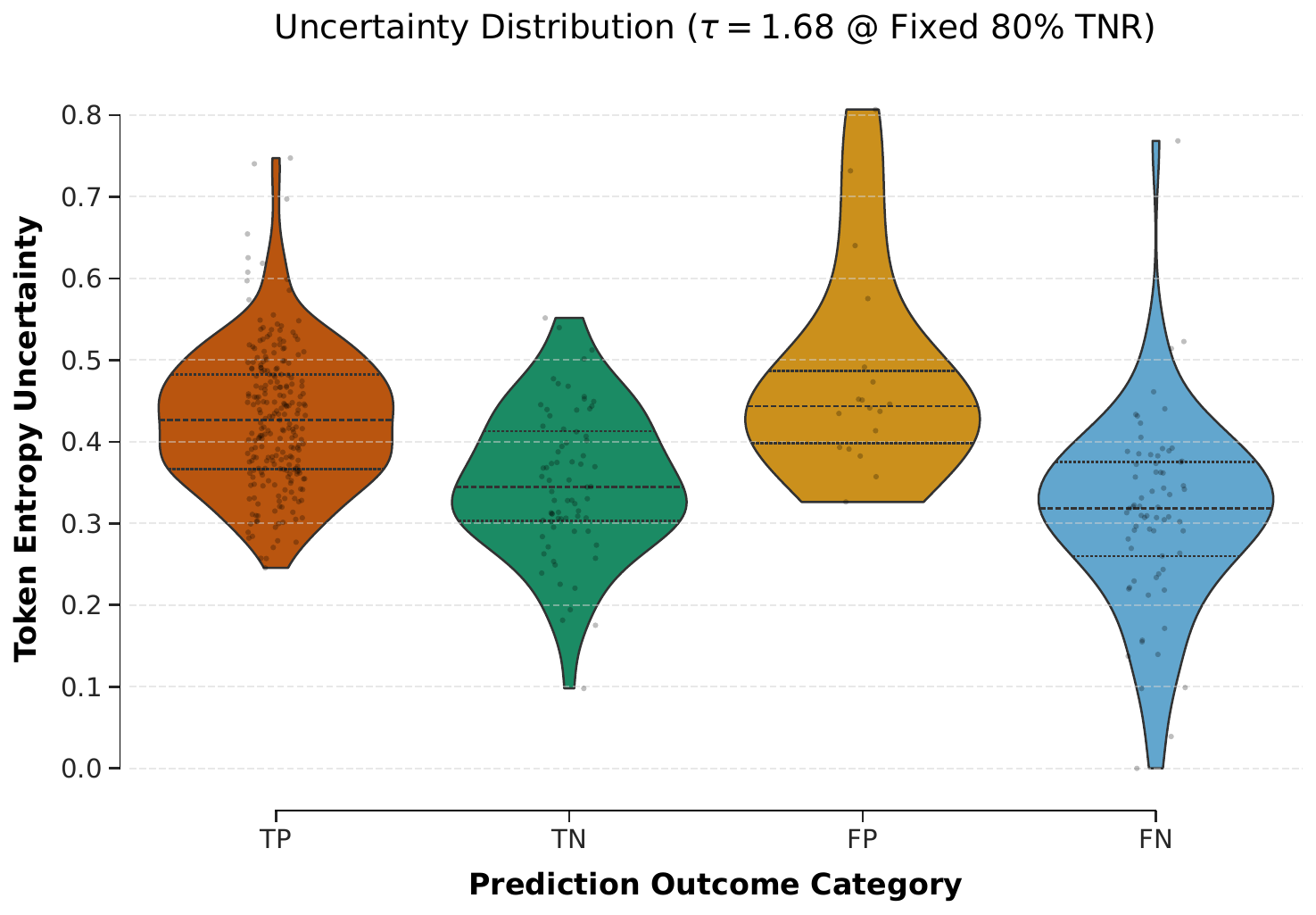}
        \caption{After Diversity Fine-Tuning}
        \label{fig:squad_dist_after}
    \end{subfigure}

    \caption{
    Token entropy distributions on the SQuAD dataset before fine-tuning, after standard supervised fine-tuning, and after diversity-oriented fine-tuning.
    }
    \label{fig:squad_distribution_all}
\end{figure*}

\begin{table}[t]
\centering
\footnotesize
\setlength{\tabcolsep}{4pt}
\renewcommand{\arraystretch}{1.1}
\begin{tabular}{lcccc@{}}
\toprule
\textbf{Models} & \textbf{TriviaQA} & \textbf{SQuAD} & \textbf{NQ-Open}  \\
\midrule
Mistral-7B-Instruct & 0.637 & 0.616 & 0.710  \\
LLaMA3-8B-Instruct  & 0.703 & 0.665 & 0.668  \\
Qwen2.5-7B-Instruct & 0.744 & 0.718 & 0.768  \\
\bottomrule
\end{tabular}
\caption{AUROC of transferring hallucination judgments from fine-tuned models to base models.}
\label{tab:transfer_auc}
\end{table}

\begin{table*}[t]
\centering
\footnotesize
\renewcommand{\arraystretch}{0.95}
\begin{tabular*}{\textwidth}{@{\extracolsep{\fill}}lcccc}
\toprule
\textbf{Method} 
& \textbf{Qwen2.5-7B} 
& \textbf{Mistral-7B} 
& \textbf{Mistral-7B} 
& \textbf{LLaMA3-8B} \\
& \textbf{NQ-Open}
& \textbf{SQuAD}
& \textbf{NQ-Open}
& \textbf{NQ-Open} \\
\midrule
Base    & 0.699 & 0.653 & 0.678 & 0.641 \\
Top-$p$ & 0.727 & 0.739 & --    & --    \\
DBS-M   & 0.686 & 0.646 & 0.664 & 0.679 \\
DBS-H   & 0.733 & 0.729 & --    & --    \\
Ours    & \textbf{0.774} & \textbf{0.765} & \textbf{0.704} & \textbf{0.714} \\
\bottomrule
\end{tabular*}
\caption{
Comparison with inference-time diversity baselines. Top-$p$ uses nucleus sampling with $p=0.95$. DBS-M and DBS-H denote moderate and high-diversity beam search, respectively.
Ours reports the best diversity-oriented fine-tuned model for each setting.}
\label{tab:decoding_baselines}
\end{table*}

\begin{table*}[t]
\centering
\footnotesize
\renewcommand{\arraystretch}{0.95}
\begin{tabular*}{\textwidth}{@{\extracolsep{\fill}}llccc|ccc}
\toprule
\textbf{Model} & \textbf{Dataset}
& \multicolumn{3}{c|}{\textbf{High-Temperature Baseline ($T=1.0$)}} 
& \multicolumn{3}{c}{\textbf{Aggressive Decoding ($T=1.2$, $N=12$)}} \\
\cmidrule(lr){3-5} \cmidrule(lr){6-8}
& 
& \textbf{Base} & \textbf{Ours} & \textbf{Gain}
& \textbf{Base} & \textbf{Ours} & \textbf{Gain} \\
\midrule
\multirow{3}{*}{Qwen}
& SQuAD    
& 0.646 & 0.810$_{\text{SFT+DPO}}$ & +0.164
& 0.689 & 0.700$_{\text{SFT}}$     & +0.011 \\
& NQ-Open  
& 0.723 & 0.774$_{\text{SFT}}$     & +0.051
& 0.676 & 0.775$_{\text{SFT+DPO}}$ & +0.099 \\
& TriviaQA 
& 0.767 & 0.769$_{\text{DPO}}$     & +0.002
& 0.726 & 0.807$_{\text{SFT+DPO}}$ & +0.081 \\
\midrule
\multirow{3}{*}{Mistral}
& SQuAD    
& 0.679 & 0.765$_{\text{SFT+DPO}}$ & +0.086
& 0.702 & 0.758$_{\text{SFT}}$     & +0.056 \\
& NQ-Open  
& 0.690 & 0.706$_{\text{DPO}}$     & +0.016
& 0.680 & 0.741$_{\text{DPO}}$     & +0.061 \\
& TriviaQA 
& 0.753 & 0.782$_{\text{DPO}}$     & +0.029
& 0.723 & 0.802$_{\text{DPO}}$     & +0.079 \\
\midrule
\multirow{3}{*}{LLaMA}
& SQuAD    
& 0.702 & 0.750$_{\text{SFT}}$     & +0.048
& 0.714 & 0.762$_{\text{DPO}}$     & +0.048 \\
& NQ-Open  
& 0.679 & 0.714$_{\text{SFT}}$     & +0.035
& 0.683 & 0.748$_{\text{DPO}}$     & +0.065 \\
& TriviaQA 
& 0.753 & 0.800$_{\text{SFT+DPO}}$ & +0.047
& 0.744 & 0.774$_{\text{SFT+DPO}}$ & +0.030 \\
\bottomrule
\end{tabular*}
\caption{
Comparison with high-temperature inference-time baselines. 
The left block reports results with un-fine-tuned models using sampling temperature $T=1.0$.
The right block reports aggressive high-diversity decoding with $T=1.2$ and $N=12$ generations.
``Ours'' reports the best diversity-oriented fine-tuned variant under the corresponding evaluation setting.
Gain is computed against the corresponding un-fine-tuned baseline.
}
\label{tab:high_temp_combined}
\end{table*}

\subsection{Comparison with Inference-Time Diversity Baselines}
\label{app:inference_baselines}

We further compare diversity-oriented fine-tuning with simpler inference-time strategies for increasing generation diversity. This experiment addresses whether the improvements of our method can be reproduced by directly increasing decoding randomness or by using diversity-oriented decoding algorithms at inference time.

We consider three types of inference-time baselines. First, we increase the sampling temperature beyond the default setting used in the main experiments. The main baseline uses high-temperature sampling with $T=0.8$ and $N=9$ sampled responses for semantic entropy computation. We further evaluate more aggressive sampling settings, including $T=1.0$ and $T=1.2$, while keeping the number of sampled generations fixed when applicable. Second, we evaluate nucleus sampling with top-$p=0.95$. Third, we evaluate diverse beam search (DBS), using both a moderate-diversity setting with 10 beams, 5 beam groups, and diversity penalty 0.5, and a high-diversity setting with 10 beams, 10 beam groups, and diversity penalty 1.0.

The results are summarized in Tables~\ref{tab:high_temp_combined} and \ref{tab:decoding_baselines}. Increasing inference-time diversity improves the baseline in some settings, confirming that semantic entropy benefits from more varied generations. However, in many cases, our fine-tuned models achieve substantially higher AUROC than the corresponding high-temperature or diversity-decoding baselines. This indicates that the gains of our method cannot be explained solely by increased sampling randomness. Instead, diversity-oriented fine-tuning changes the model's conditional generation behavior, making hallucinated cases more likely to reveal semantic inconsistency while preserving semantic stability for correct cases.

\section{Case Studies and Error Analysis}

\subsection{Additional Pilot Studies On SQuAD} 
\label{more pilot studies}

Figure \ref{fig:squad_distribution_all} presents a comparative analysis of token entropy distributions on the SQuAD dataset under three strategies: (a) Before Fine-Tuning (b) Standard SFT and (c) our proposed Diversity-Oriented Fine-Tuning. All results are evaluated at a fixed True Negative Rate of 80\%, enabling a direct comparison of uncertainty calibration across training regimes.

After Standard SFT, the entropy distributions of TP, FP, and FN show substantial overlap with the TN distribution. Notably, the TP and TN distributions are poorly separated, limiting the effectiveness of token entropy as a standalone signal for distinguishing correct responses from hallucinations. The resulting uncertainty threshold is $\tau = 1.28$.

In contrast, Diversity-Oriented Fine-Tuning further elevates overall token entropy, with the TN median rising to approximately 0.35, indicating stronger diversity effects. Crucially, this strategy preserves and improves distributional separability. As shown in Figure \ref{fig:squad_distribution_all}(c), the TP distribution is centered at a higher entropy level (around 0.45) than the TN distribution, while FP responses are also shifted toward higher uncertainty regions. This improved separation is reflected in a higher uncertainty threshold of $\tau = 1.68$, indicating that the model can accommodate greater intrinsic uncertainty while still correctly identifying 80\% of valid responses. These results demonstrate that our approach enhances expressive diversity without collapsing the uncertainty signal required for effective hallucination detection.

\subsection{Qualitative Case Studies of Diversity-Oriented Fine-Tuning} \label{case_study}

Table ~\ref{tab:case_correct} and ~\ref{tab:case_incorrect} provides a qualitative comparison illustrating the impact of fine-tuning of diversity on model generative behavior.

\begin{table*}[t]
\centering
\footnotesize
\setlength{\tabcolsep}{8pt}
\renewcommand{\arraystretch}{1.25}
\begin{tabular}{p{0.48\textwidth} p{0.48\textwidth}}
\toprule
\textbf{Before Fine-Tuning} & \textbf{After Fine-Tuning} \\
\midrule
\multicolumn{2}{l}{\textbf{Question:} Electrolysis of what can be used to produce oxygen and hydrogen?} \\
\multicolumn{2}{l}{\textbf{Reference Answer:} Water.} \\
\multicolumn{2}{l}{\textbf{Low-Temperature Answer Before Fine-Tuning:} Water.} \\
\multicolumn{2}{l}{\textbf{Low-Temperature Answer After Fine-Tuning:} Water electrolyte pair.} \\
\midrule
1. Water. & 1. Distilled water electrolysis for oxygen and hydrogen production. \\
2. Water. & 2. Water solution. \\
3. Water. & 3. Water electrolysis yields oxygen and hydrogen. \\
4. Water. & 4. Distilled water electrolysis for generating oxygen and hydrogen. \\
5. Water. & 5. Water electrolyzed. \\
6. Water. & 6. Water electrolysis for generating oxygen and hydrogen. \\
7. Water. & 7. Distilled water electrolyzed yields pure oxygen and hydrogen gases. \\
8. Water. & 8. Water molecules. \\
9. Water. & 9. Water electrolyzed. \\
\bottomrule
\end{tabular}
\caption{
Correct sample from SQuAD.
Before diversity-oriented fine-tuning, the model repeating the exact same answer. After diversity-oriented fine-tuning, the model generates diverse formulations while maintaining semantic correctness relative to the reference.
}
\label{tab:case_correct}
\end{table*}

\begin{table*}[t]
\centering
\footnotesize
\setlength{\tabcolsep}{8pt}
\renewcommand{\arraystretch}{1.25}
\begin{tabular}{p{0.48\textwidth} p{0.48\textwidth}}
\toprule
\textbf{Before Fine-Tuning} & \textbf{After Fine-Tuning} \\
\midrule
\multicolumn{2}{l}{\textbf{Question:} What color were the Broncos’ uniforms in Super Bowl 50?} \\
\multicolumn{2}{l}{\textbf{Reference Answer:} White.} \\
\multicolumn{2}{l}{\textbf{Low-Temperature Answer Before Fine-Tuning:} Orange.} \\
\multicolumn{2}{l}{\textbf{Low-Temperature Answer After Fine-Tuning:} The Broncos' attire at Super Bowl 50 was crimson.} \\
\midrule
1. Orange. & 1. Silver-and-blue scheme adopted by the Broncos for Super Bowl 50. \\
2. Orange. & 2. Deep blue jerseys and pants with white socks. \\
3. Orange. & 3. Black-and-gold Broncos attire. \\
4. Orange. & 4. Silver-and-white ensembles. \\
5. Orange. & 5. Deep blue Broncos attire. \\
6. Orange. & 6. Crimson-and-gold scheme for the event. \\
7. Orange. & 7. Silver-and-gold ensembles. \\
8. Orange. & 8. Silver-and-gold combination. \\
9. Orange. & 9. Black-and-gold Broncos attire. \\
\bottomrule
\end{tabular}
\caption{
Hallucinated sample from SQuAD.
Before diversity-oriented fine-tuning, the model repeatedly gave the same incorrect answers. After diversity-oriented fine-tuning, although the generated answers remain incorrect, they exhibit greater semantic dispersion, breaking the previously high semantic consistency and thereby improving the detectability of hallucinations.
}
\label{tab:case_incorrect}
\end{table*}

\subsection{Case Studies of Semantic Entropy on Open-Domain QA}
\label{case_study on Open-Domain QA}

This appendix presents representative NQ-Open examples to explain why semantic-entropy-based hallucination detection performs worse on open-domain QA. Since NQ-Open questions may have multiple semantically valid answers with different surface forms, high-temperature generations can remain factually plausible while exhibiting high semantic diversity. As shown in Tables \ref{tab:nqopen-mistral1}--\ref{tab:nqopen-llama2}, low-temperature answers are often consistent with the reference, but their sampled high-temperature variants may be split into multiple semantic clusters. This fragmentation inflates entropy even for non-hallucinated outputs, leading to lower AUROC on NQ-Open, especially for Mistral-7B and LLaMA3-8B.


\begin{table*}[t]
\centering
\footnotesize
\setlength{\tabcolsep}{6pt}
\renewcommand{\arraystretch}{1.25}
\begin{tabular}{p{\textwidth}}
\toprule
\textbf{Question:} what is the name of india pakistan border? \\
\textbf{Reference Answer:} International Border. \\
\textbf{Low-Temperature Answer:} The name of the India-Pakistan border is the Line of Control (LoC) in Kashmir and the International Border (IB) elsewhere. \\
\midrule
\textbf{High-Temperature Answers:} \\
1. The name of the border between India and Pakistan is the Line of Control (LoC). \\
2. The name of the India-Pakistan border is the Line of Control (LoC) in Kashmir and the International Border (IB) elsewhere. \\
3. It is called the Line of Control (LoC) or the Working Boundary (WB) in some areas. \\
4. The name of the India-Pakistan border is the Line of Control (LoC) in Kashmir and the International Border elsewhere. \\
5. The name of the India-Pakistan border is the Line of Control (LoC) in Kashmir and the International Border (IB) elsewhere. \\
6. India-Pakistan Border, also known as the Line of Control (LoC), is a de facto border that separates the disputed territory of Kashmir between India and Pakistan. \\
7. The name of the border between India and Pakistan is the Line of Control (LoC). \\
8. The name of the India-Pakistan border is the Line of Control (LoC) in Kashmir, and the International Border elsewhere. \\
9. The name of the India-Pakistan border is the Line of Control (LoC) in Kashmir and the International Border (IB) elsewhere. \\
\bottomrule
\end{tabular}
\caption{Example 1 from NQ-Open under Mistral-7B. Labeled as non-hallucination, high semantic entropy.}
\label{tab:nqopen-mistral1}
\end{table*}

\begin{table*}[t]
\centering
\footnotesize
\setlength{\tabcolsep}{6pt}
\renewcommand{\arraystretch}{1.25}
\begin{tabular}{p{\textwidth}}
\toprule
\textbf{Question:} what was the main characteristic of post-world war ii american society? \\
\textbf{Reference Answer:} Strong economic growth. \\
\textbf{Low-Temperature Answer:} Post-World War II American society was characterized by a period of economic prosperity, known as the ``Golden Age of Capitalism.'' This was due to the implementation of Keynesian economic policies. \\
\midrule
\textbf{High-Temperature Answers:} \\
1. After World War II, American society underwent significant changes, including economic growth, suburbanization, and the rise of consumer culture. \\
2. After World War II, American society experienced a period of prosperity and growth, often referred to as the ``Golden Age.'' \\
3. Post-World War II American society was characterized by economic prosperity, cultural expansion, and political conservatism. \\
4. The main characteristic of post-World War II American society was a period of economic prosperity, known as the ``Golden Age of Capitalism,'' due to the expansion of consumer goods, the growth of the suburbs, and the rise of the middle class. \\
5. Post-World War II American society was characterized by a period of economic growth, known as the post-war economic boom. \\
6. After World War II, American society experienced a period of prosperity and growth known as the postwar economic boom. \\
7. The main characteristic of post-World War II American society was a period of economic prosperity known as the ``Golden Age of Capitalism.'' This was characterized by high economic growth, low unemployment, and rising standards of living for many Americans. \\
8. Post-World War II American society was characterized by a period of economic growth and prosperity known as the post-war economic boom. \\
9. Post-World War II American society was characterized by a period of economic prosperity and growth, known as the post-war economic boom, driven by factors such as returning soldiers starting families. \\
\bottomrule
\end{tabular}
\caption{Example 2 from NQ-Open under Mistral-7B. Labeled as non-hallucination, high semantic entropy.}
\label{tab:nqopen-mistral2}
\end{table*}


\begin{table*}[t]
\centering
\footnotesize
\setlength{\tabcolsep}{6pt}
\renewcommand{\arraystretch}{1.25}
\begin{tabular}{p{\textwidth}}
\toprule
\textbf{Question:} where does hydrogen peroxide come from in the body? \\
\textbf{Reference Answer:} Nearly all living cells. \\
\textbf{Low-Temperature Answer:} Hydrogen peroxide is produced within the body's cells. \\
\midrule
\textbf{High-Temperature Answers:} \\
1. Hydrogen peroxide is produced in the body through the action of enzymes. \\
2. Hydrogen peroxide can be found in the body, specifically in the neutrophils. \\
3. Hydrogen peroxide is produced within the body's neutrophils. \\
4. Hydrogen peroxide is produced within the body's neutrophils. \\
5. The phagocytes. \\
6. The reaction between oxygen and hydrogen in the body. \\
7. The body generates hydrogen peroxide through the process of phagocytosis. \\
8. Certain cells. \\
9. Catalase. \\
\bottomrule
\end{tabular}
\caption{Example 1 from NQ-Open under LLaMA3-8B. Labeled as non-hallucination, high semantic entropy.}
\label{tab:nqopen-llama1}
\end{table*}

\begin{table*}[t]
\centering
\footnotesize
\setlength{\tabcolsep}{6pt}
\renewcommand{\arraystretch}{1.25}
\begin{tabular}{p{\textwidth}}
\toprule
\textbf{Question:} What was one reason south carolina gave for its decision to secede from the union? \\
\textbf{Reference Answer:} To preserve slavery. \\
\textbf{Low-Temperature Answer:} Preservation of slavery. \\
\midrule
\textbf{High-Temperature Answers:} \\
1. Preservation of slavery. \\
2. Preservation of slavery. \\
3. States' rights. \\
4. The threat to its right to own slaves. \\
5. Preservation of slavery. \\
6. The issue of slavery. \\
7. The preservation of slavery. \\
8. States' rights. \\
9. Preservation of states' rights. \\
\bottomrule
\end{tabular}
\caption{Example 2 from NQ-Open under LLaMA3-8B. Labeled as non-hallucination, high semantic entropy.}
\label{tab:nqopen-llama2}
\end{table*}

\end{document}